# Predicting recovery following stroke: deep learning, multimodal data and feature selection using explainable AI

Adam White[1], Margarita Saranti[2], Artur d'Avila Garcez[1], Thomas M. H. Hope[4], Cathy J. Price[4], Howard Bowman[2,3*]

1.      Department of Computer Science, City, University of London, UK
2.      School of Psychology, University of Birmingham, UK
3.      School of Computing, University of Kent, UK
4.      Wellcome Centre for Human Neuroimaging, University College London, UK

* Corresponding author at: School of Psychology, University of Birmingham, Edgbaston, Birmingham, B15 2TT, UK

## Abstract

Machine learning offers great potential for automated prediction of post-stroke symptoms and their response to rehabilitation.  Major challenges for this endeavour include the very high dimensionality of neuroimaging data, the relatively small size of the datasets available for learning and interpreting the predictive features, and how to effectively combine neuroimaging and tabular data (e.g. demographic information and clinical characteristics). This paper evaluates several solutions based on two strategies. The first is to use 2D images that summarise MRI scans. The second is to select key features that improve classification accuracy. Additionally, we introduce the novel approach of training a convolutional neural network (CNN) on images that combine regions-of-interest (ROIs) extracted from MRIs, with symbolic representations of tabular data.

We evaluate a series of CNN architectures (both 2D and a 3D) that are trained on different representations of MRI and tabular data, to predict whether a composite measure of post-stroke spoken picture description ability is in the aphasic or non-aphasic range.  MRI and tabular data were acquired from 758 English speaking stroke survivors who participated in the PLORAS study.  Each participant was assigned to one of five different groups that were matched for initial severity of symptoms, recovery time, left lesion size and spoken description scores months or years post-stroke. Training and validation was carried out on the first four groups. The fifth (locked box) group was used to test how well model accuracy generalises to new (unseen) data.

The classification accuracy for a baseline logistic regression was 0.678 for lesion size alone, rising to 0.757 and 0.813 when initial symptom severity and recovery time were successively added.  The highest classification accuracy (0.854), area under the curve (0.899) and F1 (0.901) were observed when 8 regions of interest was extracted from each MRI scan and combined with lesion size, initial severity and recovery time in a 2D Residual Neural Network.  This was also the best model when data were limited to the 286 participants with moderate or severe initial aphasia (area under curve = 0.864), a group that would be considered more difficult to classify.

Our findings demonstrate how imaging and tabular data can be combined for high post-stroke classification accuracy, even when the dataset is small in machine learning terms.  We conclude by proposing how the current models could be improved to achieve even higher levels of accuracy using images from hospital scanners.

## Introduction

Modern healthcare has become good at keeping patients alive following a stroke. Consequently, there are increasingly many stroke-survivors with debilitating impairments that they may live with for many years. Of impairments following stroke, language deficits can be particularly distressing, since they limit the ability to communicate with others, impacting relationships with friends and family, as well as work opportunities. Accordingly, post-stroke rehabilitation is critically important.

Targeted therapy for post-stroke aphasia has been shown to bring benefit, even in the chronic stage (Menahemi-Falkov et al., 2021; Pierce, 2023). Ideally, one would like to predict deficits soon after stroke and use that information to target rehabilitation at the identified deficit. Furthermore, one would like prediction of deficits to be obtained automatically, or at least with the assistance of modern machine learning.

Modern AI, through its focus on deep learning, offers great potential for automated prediction (Roohani et al., 2018; Chauhan et al. 2019). However, although there is a determined effort to acquire large datasets in stroke research, they remain small in machine learning terms. This means that the signal-to-noise level is relatively low, and this has the consequence that feature selection is likely to be needed. For example, a high resolution T1 weighted MRI scan has hundreds of thousands of voxels/ features, and the number of trainable parameters in a 3D convolutional neural network (CNN) is in the millions. Yet the number of patients in the stroke datasets rarely exceeds the low thousands. This is very little data to train such high dimensionality. This paper proposes and evaluates two possible strategies. The first is to use 2D images that summarise MRI scans. The second is to identify key features to be added to image processing that can lead to good classifications.

A further challenge is how to combine MRI data with tabular data (i.e. demographic information and clinical characteristics). As we will illustrate, there has only been limited success in developing multimodal deep learning systems that combine MRI and tabular data. But, as discussed below, there is robust evidence that both MRI and tabular data have value in predicting post-stroke language deficits.

In this paper, our objectives are to,

1) provide a state-of-the-art assessment of the effectiveness of deep learning when predicting a functionally informative measure of language deficits for stroke survivors, who were assessed months or years post-stroke (spoken picture description);
2) assess the value of multimodal deep learning models, which include both images and tabular data; and
3) determine the key "information-bearing" feature dimensions.

We introduce our novel approach for training CNNs on images that combine regions-of-interest (ROIs) extracted from MRIs, with symbolic representations of tabular data.

Experiments were carried out with a series of CNN architectures (both 2D and a 3D) that combined MRI and tabular data to predict whether spoken picture description scores were in the aphasic or non-aphasic range. Several of our experiments used a Residual Neural Network (ResNet) model, as this type of CNN provides state of the art levels of accuracy in medical imaging, due to its "skip connections" enabling the scaling-up to large numbers of layers. There are a variety of 2D and 3D ResNet models, typically labelled with a number following "ResNet" (e.g. ResNet-18) that refers to the number of layers in the model.

All analyses were carried out using MRI and tabular data from the Predicting Language Outcome and Recovery After Stroke (PLORAS) database (Seghier et al., 2016). This includes patients' high resolution T1-weighted structural MRI brain scans that are acquired months or years post stroke, lesion images derived from the MRIs, and tabular data including language and cognitive scores from the Comprehensive Aphasia Test (CAT) battery (Swinburn, et al., 2004). PLORAS excludes patients with evidence of other neurological conditions. To ensure that low language scores were not a consequence of non-stroke related language proficiency, we also excluded patients whose native language was not English.

Hope et al. (2013) employed Gaussian process regression models to predict the CAT spoken picture description scores that are also of interest in the current study. A baseline model using just demographic data and elapsed time since stroke gave an R-squared of 0, using data from 270 patients from the PLORAS database (Seghier et al., 2016). The R-squared was increased to 0.33 when adding lesion volume; and 0.59 when adding lesion loads that indicate the proportions of anatomically defined grey and white matter regions of interest (ROI) that are categorised as "lesioned" in each patient.

Hope et al. (2018) analysed whether disrupted white matter connectivity adds unique prognostic information for post-stroke aphasia recovery. Baseline regression models were trained using the PLORAS data of 818 patients, including demographic data, elapsed time since stroke, lesion volume and lesion loads of grey matter ROIs, where lesion load was the proportion of each ROI damaged in each binary lesion image. The baseline models were then compared to a series of models that added or replaced the data from the baseline model with white matter connectivity data. The best Pearson R scores reported for the spoken description score were 0.73. Overall, it was found that adding connectivity data did not improve prediction accuracy for patient language skills, a finding that was also observed in an independent dataset by Zhao et al. (2023). Hope et al. emphasise that their findings do not exclude white matter disruption being a key casual mechanism for post-stroke cognitive symptoms. This is because lesions may result in highly correlated grey matter and white matter damage. Hence grey matter damage could be a suitable proxy in prognostic models, even if white matter damage is etiologically important.

Roohani et al. (2018) trained a CNN using 2-D stitched images created from 1,211 PLORAS MRI scans. Each image consisted of sixty-four axial cross-sectional slices from each MRI scan (Figure 1, left). The slices were always stitched in the same order, so that a voxel location in the stitched images always corresponded to the same brain location. Roohani et al. motivated their stitched image format on the grounds that there was insufficient data to effectively train a 3D network. By contrast, using 2D stitched images reduces the number of trainable parameters, whilst still capturing contextual information across scans. The CNN achieved a prediction accuracy of 79% for classifying patients' spoken picture description scores (aphasic or not aphasic), based on a threshold score of 60. A second analysis was carried out by combining the feature vector from the final layer of the convolutional layer with demographic data, and then regressing against spoken description scores; giving an R-squared of 0.6. Roohani et al.'s analysis suggests that the stitched image format successfully captures the predictive signal within an MRI scan, however it is not directly comparable with either of the Hope et al. (2013, 2018) papers, as each uses a different subset of participants from the PLORAS database.

Chauhan et al. (2019) compared the performance of a 3D CNN trained on post-stroke MRI scans with both a ridge regression and a support vector regression model trained on features of lesion images extracted by principal component analysis. A hybrid model was also trained that combined the lesion image features with features extracted from the 3D CNN. This was carried out with data from 98 patients with language deficits from a Washington University School of Medicine dataset. The support vector regression had the highest R-squared of 0.66 compared to the 3D CNN's 0.63.

There are very few published multimodal CNNs that combine MRI data with tabular clinical data. We are unaware of any papers using multimodal CNNs for predicting language outcomes after stroke, however there are several papers on diagnosing Alzheimer's that are relevant. Esmaeilzadeh et al. (2018) and Liu et al. (2019) both use 'Early Fusion' models (Huang et al., 2020). Early Fusion models consist of a CNN that learns a latent representation of the input images, the latent representation is then concatenated with the tabular data, before being passed through some fully connected layers. The Liu et al. (2019) model first identifies discriminative anatomical landmarks from MRI images, extracts image patches around these landmarks and passes these patches into a CNN. The feature maps from the last convolutional layer are then concatenated with demographic data before being passed through some fully connected layers. Wolf et al. (2022) criticise such approaches as failing to enable fine grained interaction between voxels and tabular data. They propose a multimodal 3D convolution neural network called Dynamic Affine Feature Map Transform (DAFT). DAFT employs a modified 3D ResNet architecture in which tabular data scales and shifts the feature maps of the ResNet's final layer. Wolf et al. trained DAFT on MRI and tabular data for diagnosing and predicting Alzheimer's disease. In their experiments DAFT had higher balanced accuracy, AUC and F1 scores than either a baseline linear regression, an Early Fusion model, or a 3D ResNet.

The remainder of the paper is organised as follows. Section 2 specifies the data that was extracted from the PLORAS dataset, and how this was used to create new image datasets displaying features such as ROIs and symbolic representations of tabular data. A summary is also provided of an explainable AI method called CLEAR Image that was used to identify key ROIs. Section 3 specifies the experiments that were carried out using a variety of CNN architectures. It also explains how the project avoided overfitting by employing a strategy of cross-validation with a hold out 'lock box' (Hosseini et al., 2020). The results are presented in Section 4, highlighting the potential of using images that combine ROIs and symbolic representations of tabular data. Section 5 discusses our findings and indicates directions for future work. Section 6 identifies key limitations, including that the MRI data was restricted to research quality scanners. Section 7 concludes the paper.

2. ***Methods and Materials***

2.1 *Dataset*

The participants were 758 stroke survivors from the PLORAS database (Seghier et al., 2016) The male to female ratio was 2.3:1, and the average age at stroke was 56.1. The dataset used for the current study consists of MRI scans, their associated tabular data and six two-dimensional image datasets that are derived from the MRI scans (see subsections 2.1.1 to 2.1.6). Three PLORAS tabular features were identified *a priori* as being of prognostic relevance to recovery from aphasia: (i) Initial severity of aphasia after stroke (henceforth: initial severity), see Lazar et al. (2010), Benghanem et al. (2019) (ii) Left hemisphere lesion size (henceforth: left lesion size), see Hope et al. (2013), Thye & Mirman (2018), Benghanem et al. (2019) (iii) Recovery time - which is defined as the time between the stroke and the CAT tests, see Hope et al. (2013), Johnson et al. (2022).

In this paper, initial severity was assessed by patient report (as in Roberts et al., 2022). A patient was classified as *severe* if they were conscious, physically capable of attempting to speak but unable to speak due to aphasia; *moderate* if they were able to produce words but not sentences; *mild* if they could produce lexically meaningful short sentences and *normal* if they did not report an impairment. There is an additional category for patients who were either unconscious and hence could not be tested, or whose score was missing. Initial severity was treated as a categorical rather than ordinal measure because the unconscious/missing values cannot be ranked relative to the other values. Initial severity scores were distributed: 25.5% severe, 12.3% moderate, 24.8% mild, 17.2% normal and 20.2% unconscious or missing.

The outcome of interest for this paper was the total score from the CAT spoken picture description task. This task requires participants to conceptualise events in a scene, retrieve the words associated with the objects and actions, formulate sentences, and generate the associated speech sounds. It objectively measures the building blocks of connected speech, including the number and appropriateness of information carrying words, syntactic variety, speed ratings and grammatical accuracy. We focused on predicting the overall score which provides a reasonable proxy for participants' language skills in more naturalistic contexts. The overall scores were standardised into T-scores (not to be confused with the t-statistic) that measure patient performance relative to an independent sample of participants without aphasia. In the current paper, we classified scores that were less than 60 as aphasic, as this is rarely observed in participants from the PLORAS database who do not have any identifiable brain damage. The distribution of spoken picture description scores was skewed, with 34% having a score less than 60 (i.e. in the aphasic range). For patients with severe or moderate initial severity scores, 44.5% had spoken description scores less than 60.

The MRI scans, from our 758 participants were acquired by research-dedicated 3T scanners between 30th June 2010 and 14th March 2020 (when data collection was stopped by Covid-19 restrictions). Participants recruited prior to these dates were not included because initial severity scores were not routinely collected. The MRIs were spatially normalised into standard Montreal Neurological Institute space using the unified segmentation approach (Crinion et al., 2007). Lesion segmentation used the automated lesion identification (ALI) approach, specified by Seghier et al. (2008), which outputs a 3D whole brain binary lesion image (lesioned or not lesioned) for each patient's brain. Left lesion size was an estimate of the number of damaged voxels in each patient's left hemisphere binary lesion image (see Hope et al., 2013).

The project dataset was partitioned into five groups, such that each group was balanced in terms of recovery time, initial severity, left lesion size and spoken description score. All this paper's training and validation was carried out on the first four groups, with the fifth group being held back as a lock box. A lock box is a subset of the dataset removed from the analysis pipeline before any optimisation begins, and not accessed until after all hyperparameter adjustments and training is completed. As long as no decisions concerning the set-up or training of data is made on the lock box, which would be the case if accuracy on the lock box is only assessed once, performance on the lock box is a fair test of generalization (Hossenini et al., 2020).

*2.1.1 Stitched MRI dataset*

The 2D stitched MRI used in this project were produced to the same specification as used by Roohani et al. (2018). These images do not rely on any lesion segmentation processing. They are created by displaying sixty-four axial cross-sectional spatially normalised MRI slices in a single 2D 632 x 760 image (see Figure 1, left). These 2D images are then down-sampled to 256 x 256 as part of preprocessing for the CNNs. The down-sampling leads to some distortion in the shapes of the MRI slices and also some loss of information. The degree to which the resulting images can still be used to generate accurate forecasts was one of the questions for the experiments performed for this paper.

*2.1.2 Grey Matter Regions of Interest (GM-ROI) dataset*

The original 2D stitched MRI images (prior to down-sampling) were parcellated into grey and white matter anatomical ROIs. The grey matter (GM) ROIs are described here, and the white matter (WM) ROIs are described in Section 2.1.5. The GM-ROIs were defined by the Automatic Anatomical Labelling atlas (Tzourio-Mazoyer et al., 2002). Those that contributed most to the ResNet-18's predictions (see below) were considered "key anatomical ROIs" and stitched together into "GM-ROI images" (Figure 1, right). A possible advantage of using ROI images is that the ROIs can be kept at the original resolution of the MRI scan and hence no information is lost, whereas the stitched MRI were down-sampled (as described above). Using GM-ROI images may also reduce the risk of the curse of dimensionality (Altman, N., Krzywinski, M., 2018) compared to the stitched MRI dataset. There are also *a priori* grounds for believing that the 2D GM-ROI images reduce redundant dimensions. For example, the most relevant ROIs for aphasia are known to be in the left hemisphere. Furthermore, there can be a significant degree of duplication in the predictive information contained within an MRI slice, as a lesion that causes aphasia is likely to damage multiple ROIs, including some that are functionally irrelevant to aphasia (Seghier & Price, 2023).

The key ROIs were identified by first training a ResNet-18 neural network on the original stitched MRI dataset to predict spoken description scores >= 60 (i.e. full recovery). An explainable AI method called CLEAR Image (see subsection 2.2) then identified which of the 116 GM-ROIs were most important to the ResNet-18's predictions. CLEAR Image analysed 100 predictions made by the ResNet-18 and calculated each GM-ROI's average feature importance score. The key GM-ROIs (all left hemisphere) were, in order of importance: (i) superior temporal gyrus, (ii) middle temporal gyrus, (iii) inferior frontal gyrus - triangular, (iv) postcentral gyrus, (v) supramarginal gyrus, (vi) inferior frontal gyrus - opercular, (vii) insula gyrus, (viii) caudate gyrus, (ix) temporal pole, (x) inferior parietal, (xi) middle frontal gyrus, and (xii) hippocampus gyrus. Four of these are temporal lobe regions (i, ii, ix, xii), three are parietal lobe regions (v, x, iv), and three are front lobe regions (iii, vi, xi).

Cross-validation was then used to determine the number of GM-ROIs to include in the GM-ROI images. For example, if the number is three then the images will display the three highest scoring GM-ROIs, i.e. left superior temporal gyrus, middle temporal gyrus, inferior frontal gyrus-triangular. The cross-validation was over all permutations of the 3 learning rates (see below) and number of GM-ROIs to include which ranged from 3 to 12. It was found that the top eight ROIs minimised the ResNet-18's loss function (see Figure 2). The use of only one level of cross-validation to fit hyper-parameters, such as number of GM-ROIs, could induce over-fitting; however, our use of a lock box allows us to test whether our ultimate quantification of overall accuracy generalises well to new data, at least with the variability inherent to the PLORAS dataset (Hosseini et al., 2020).

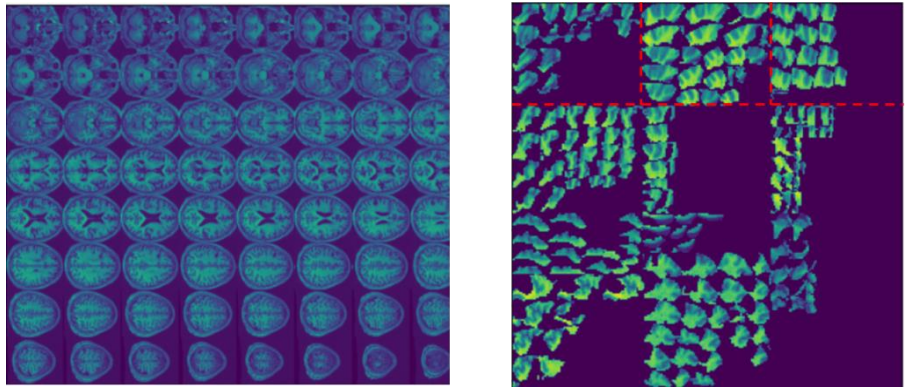

Figure 1. Left: An example of a stitched MRI consisting of sixty-four axial cross-sectional slices from an MRI scan. Right: A GM-ROI Image consisting of the 12 key (most predictive) GM-ROIs (see Section 3.2). The dotted red lines have been added to this figure for visual clarity, demarcating the boundaries of the left superior temporal gyrus, middle temporal gyrus and inferior frontal gyrus-triangular.

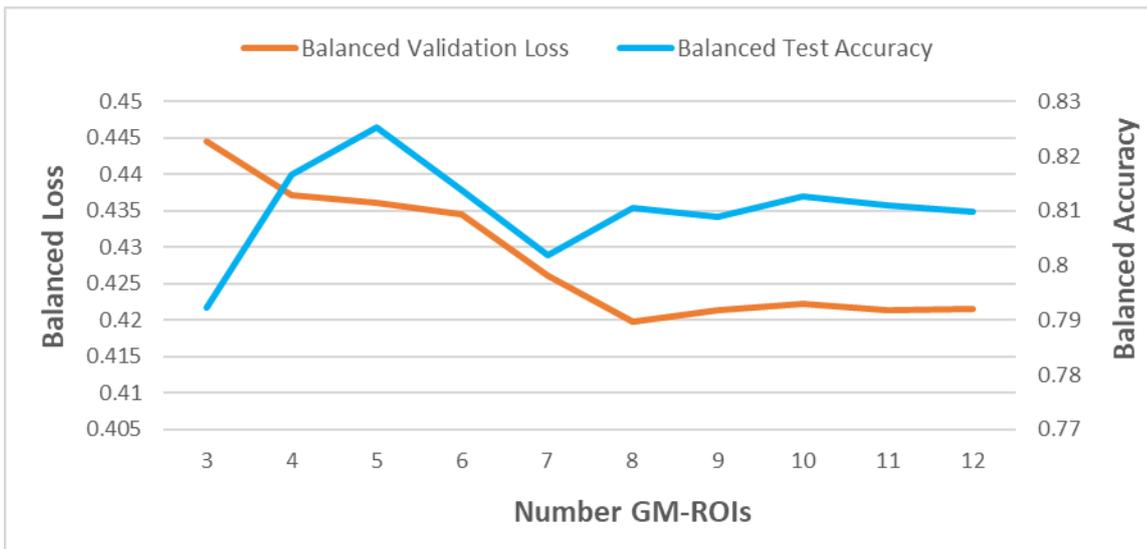

Figure 2. Plots of how balanced validation loss and balanced test accuracy vary with the number of GM-ROIs displayed in GM-ROI Images. The balanced validation loss was used to determine that 8 GM-ROIs should be included in each GM-ROI Image (selecting the number of GM-ROIs based on highest balanced test accuracy would be an example of overfitting). Note that the balanced test accuracy only varies slightly with number of GM-ROIs, achieving >0.79 with only three GM-ROIs.

### 2.1.3 Hybrid Stitched MRI dataset

Hybrid stitched MRI combine the stitched MRI with symbolic representations of initial severity, lesion size and recovery time. The choice of symbols and how to represent a feature's values was largely arbitrary, the only criteria being that the neural networks to be trained following the addition of the symbols should be sensitive to these representations. Left lesion size is a continuous feature and was represented by a pentagon symbol whose radius varies in proportion to its value. Recovery time was represented by a pie-slice of fixed size whose intensity varies in proportion to its value. Each initial severity category was represented by a different symbol, for example moderate by a triangle, normal by an ellipse, and unconscious / missing by a star. In order to create space for the tabular features in the hybrid stitched MRI dataset, four MRI slices were removed (see Figure 3., left). The excluded slices being the four most dorsal, which are rarely lesioned in our dataset.

*2.1.4 Hybrid GM-ROI dataset*

Hybrid ROI images combine the GM-ROIs and the three tabular features (see Figure 3, right). The number of GM-ROIs displayed in each image was determined using the cross-validation process described above for the ROI dataset, but with the images now also including the three tabular features. This identified that the seven top GM-ROIs should be included.

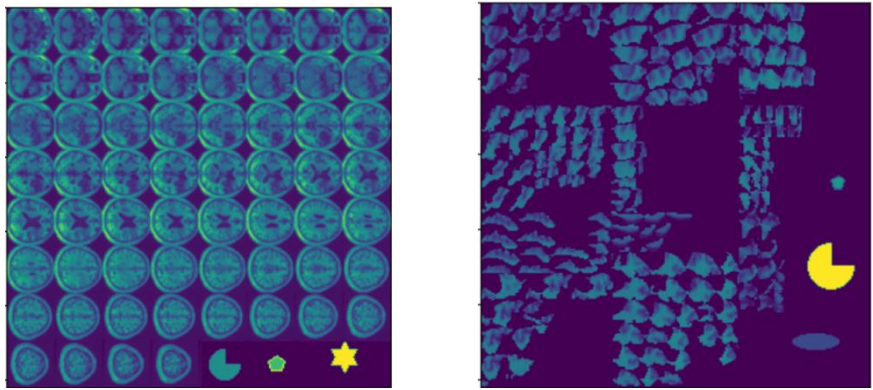

Figure 3. Left: A Hybrid stitched MRI, after pre-processing which reshapes it to 256 x 256. Right: A Hybrid ROI image consisting of the top seven ROIs plus the symbols for initial severity (normal for this patient), left lesion size and recovery time.

*2.1.5 White Matter Regions of Interest (WM-ROI) dataset*

The NetBrainLab Atlas (www.natbrainlab.co.uk/atlas-maps) was used to parcellate each of the stitched images. The voxels corresponding to three key white matter tracts were then copied from each stitched image to create new White Matter Tracts images. The three tracts were the: arcuate, internal capsule and inferior longitudinal fasciculi.

*2.1.6 Hybrid White Matter Tracts dataset*

Hybrid White Matter Tract images combine the three white matter tracts with the three tabular features.

*2.2. CLEAR Image*

CLEAR Image (White et al., 2023) was used to identify which of the 116 GM-ROIs were most salient to an image's classification probability. CLEAR Image is a perturbation-based explainable AI method that was enhanced for this paper to use brain atlases, and also contrast MRI images. The key idea behind perturbation methods is to parcellate an image into ROIs, perturb the ROIs, and then determine how much each ROI affects a neural network's classification probability. Consider an example where a neural network has assigned a stitched MRI, *S*, a classification probability of 0.96. CLEAR Image creates a perturbed image *S'* by replacing an ROI of image *S* with the same ROI taken from a 'contrast' *S''* selected from a stitched MRI with a low predicted classification. CLEAR Image then passes the perturbed image *S'* through the neural network and records how much the classification probability changes. By creating a large number (>1000) of perturbed images in which different combinations of ROIs are replaced and the changes in classification probability recorded, CLEAR Image creates a regression dataset. A logistic regression is then performed, whose coefficients give the feature importance score for each ROI. An example of a CLEAR Image explanation is shown in Figure 4. For a full specification of the CLEAR Image method and a comparison with other perturbation methods see White et al. (2023, 2021).

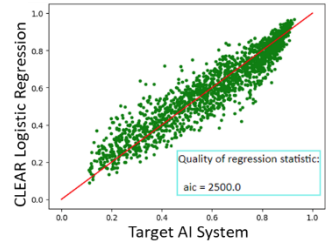
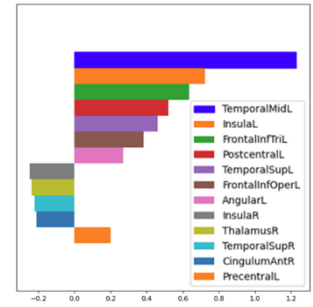
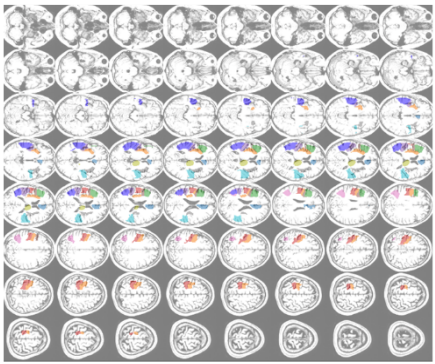
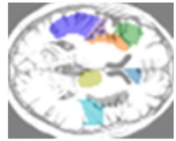

Figure 4. Example of a Clear Image output. This explains the classification probability determined by ResNet-18 from the stitched MRI of patient 108. CLEAR Image estimates the feature importance scores that the ResNet-18 has implicitly used in determining the classification probability. CLEAR Image also shows the logistic regression equation it generated for this stitched MRI, some counterfactuals and fidelity errors - these are explained in White et al. (2023).

## 3.  Experiments

Twelve sets of experiments were initially carried out:

1. A baseline logistic regression.
2. ResNet-18 fine-tuned on the Stitched MRI dataset.
3. A lightweight CNN, based on the 2D CNN used by Roohani et al. (2018), trained on the Stitched MRI dataset.
4. 3D ResNet10 fine-tuned on MRI scans dataset.
5. Early fusion model trained with the Stitched MRI dataset and tabular data.
6. Dynamic Affine Feature Transform – a multimodal 3D CNN trained on the MRI scans dataset and tabular data.
7. ResNet-18 fine-tuned on GM-ROI dataset.
8. ResNet-18 fine-tuned on WM-ROI dataset.
9. ResNet-18 fine-tuned on the Hybrid Stitched MRI dataset.
10. A lightweight CNN trained on the Hybrid Stitched MRI dataset.
11. ResNet-18 fine-tuned on the Hybrid GM-ROI dataset.
12. ResNet-18 fine-tuned on the Hybrid WM-ROI dataset.

Cross-validation was used to train each of the neural networks, with the held back test dataset (i.e. the lock box) being used to determine (final) test accuracy. The neural networks were trained for a maximum of 200 epochs using an early stopping rule that selected the epoch with the minimum class-weighted binary cross-entropy validation loss. The cross-validation selected from three learning rates: 1e-4, 5e-4, 1e-5. The neural networks were trained using stochastic gradient descent, with the exception of the lightweight neural network, which (following Roohani et al., 2018) used Root Mean Square Propagation. All neural networks used a single parameter variant of Platt scaling to calibrate the classification probabilities (Guo et al., 2017). The ResNet-18 models used ImageNet pretrained weights but were fine-tuned as part of the cross validations. Each experiment was repeated for twenty random number seeds.

The evaluation metrics were unbalanced accuracy of the predictions, balanced accuracy of predictions (i.e. the average of sensitivity and specificity), area under the ROC Curve and F1-score (i.e. the harmonic mean of precision and sensitivity). We also report the same four metrics for patients with initial severity scores that are severe or moderate, as these were taken as being clinically the most difficult to predict; see (Bowman et al.,2021; Hope et al., 2019; Bonkhoff et al., 2020) for a discussion of why patients at ceiling can inflate estimates of recovery performance. The relative importance of balanced and unbalanced accuracies was a question of interest. Balanced accuracy is perhaps the easiest to interpret. However, if the proportion of patients with and without spoken-picture description deficits in our dataset reflects the presentation of patients at hospital stroke units, then unbalanced accuracy would be more representative of the effectiveness of our machine learning classification, since it would reflect the prior probability of patients presenting with particular conditions. Accordingly, unbalanced accuracy might also be a relevant measure to consider.

Additional information on some of the experiments is provided in subsections 3.1 to 3.5.

*3.1    Baseline regression*

A binary logistic regression model was created to provide a baseline forecast for the paper. Its independent variables were the three *a priori* features: left lesion size, initial severity, recovery time. Cross-validation was not used for the logistic regression, allowing all four groups to be used for training (but excluding the fifth group, i.e. the lock box).

*3.2    Lightweight neural network trained on Stitched MRI*

This is the neural network used by Roohani et al. (2018). The basic building block was the commonly used sequence of a 2D convolution, followed by a ReLU function and a max pooling function.  This was repeated six times.  Such shallow architectures have been suggested to perform similarly to deeper networks such as ResNet-50 and Inception v-3 when applied to medical images (Raghu et al., 2019).

*3.3    3D ResNet10 trained on MRI scans*

MED3D's ResNet10 and ResNet-18 were evaluated. These were pretrained using the 3DSeg-8 dataset, which was aggregated from several medical challenges (Chen et al., 2019).

*3.4    Multimodal trained on both stitched MRI slices and tabular data*

An early fusion model using a ResNet-18 with the Stitched MRI dataset. The feature maps from the last convolutional layer of a ResNet-18 are concatenated with the three *a priori* tabular features and then passed through a fully connected layer.

*3.5    Dynamic Affine Feature Transform (DAFT)*

We use the same DAFT-Resnet model that Wolf et al. applied in their Alzheimer's study. Their modified ResNet is lightweight, with its four blocks having 4, 8, 16 and 32 output channels respectively.

4.    **Results**

Accuracy results are shown in Table 1. All the models performed well with balanced accuracies (all patients) exceeding 0.800.  For the logistic regression (without stitched or ROI images), all three tabular features were statistically significant ($p < 0.001$) although, these significant findings are, to some extent, carried by the very high degrees of freedom associated with these tests (Lorca-Puls et al., 2018). If the logistic regression was run just with 'left lesion size' as the independent variable, the balanced accuracies dropped to 0.678/0.694 (all patients/patients with initial severity severe and moderate). Adding 'initial severity' increased the balanced accuracies to 0.757/0.706, further adding 'recovery time' gave balanced accuracies of 0.813/ 0.780. The logistic regression results do not have confidence intervals, as the regressions used Statsmodel's deterministic Broyden–Fletcher–Goldfarb–Shanno optimization method (www.statsmodels.org/stable/optimization.html) and hence the results did not vary with random seed.

The strongest results came from using the ResNet-18 with the Hybrid ROI images; improving balanced accuracy by approximately 0.04 compared to the baseline logistic regression. The models trained with the Stitched MRI dataset performed similarly to the ResNet3D, suggesting that the 2D images retained the key prognostic information contained in the 3D scans. The models trained on images referencing only grey matter (GM-ROI and Hybrid GM-ROI) outperformed their corresponding models trained only on white matter images (WM-ROI and Hybrid WM-ROI). The unbalanced accuracy and balanced accuracy results for severe/moderate initial severity are almost identical due to the test dataset being approximately balanced for these two groups.

|  | I | T | All Patients | | Initial Severity: Severe or Moderate | |
| --- | --- | --- | --- | --- | --- | --- |
|  |  |  | Accuracy | Balanced accuracy | Accuracy | Balanced accuracy |
| Logistic regression | X | Y | 0.847 | 0.813 | 0.782 | 0.780 |
| Stitched MRI w/ ResNet-18 | Y | X | 0.823 +/- 0.003 | 0.807 +/- 0.004 | 0.746 +/- 0.007 | 0.746 +/- 0.007 |
| Stitched MRI w/ Lightweight CNN | Y | X | 0.825 +/- 0.003 | 0.801 +/- 0.004 | 0.739 +/- 0.006 | 0.739 +/- 0.006 |
| MRI Scans w/ ResNet3D | Y | X | 0.818 +/- 0.005 | 0.805 +/- 0.005 | 0.747 +/- 0.010 | 0.747 +/- 0.010 |
| Early Fusion Hybrid w/ ResNet-18 | Y | Y | 0.820 +/- 0.003 | 0.800 +/- 0.003 | 0.732 +/- 0.006 | 0.732 +/- 0.007 |
| DAFT | Y | Y | 0.818 +/- 0.004 | 0.814 +/- 0.004 | 0.758 +/- 0.011 | 0.759 +/- 0.010 |
| Hybrid GM-ROIs w/ ResNet-18 | Y | Y | **0.866 +/- 0.002** | **0.854 +/- 0.002** | **0.820 +/- 0.004** | **0.821 +/- 0.004** |
| Hybrid WM-ROI w/ ResNet-18 | Y | Y | 0.845 +/- 0.002 | 0.834 +/- 0.003 | 0.784 +/- 0.004 | 0.784 +/- 0.004 |
| Hybrid Stitched MRI w/ ResNet-18 | Y | Y | 0.838 +/- 0.003 | 0.829 +/- 0.004 | 0.771 +/- 0.006 | 0.771 +/- 0.006 |
| Hybrid Stitched Image w/ Lightweight CNN | Y | Y | 0.829 +/- 0.003 | 0.819 +/- 0.003 | 0.762 +/- 0.004 | 0.763 +/- 0.005 |
| GM-ROIs w/ ResNet-18 | Y | X | 0.832 +/- 0.003 | 0.811 +/- 0.004 | 0.764 +/- 0.008 | 0.763 +/- 0.008 |
| WM-ROI w/ ResNet-18 | Y | X | 0.818 +/- 0.002 | 0.803 +/- 0.003 | 0.741 +/- 0.006 | 0.741 +/- 0.006 |

Table 1. Accuracy results on the test data. (I = uses image data, T = uses tabular data.)

The area under the ROC curve (AUC) and F1 scores are shown in Table 2. As seen in the balanced and unbalanced accuracies (Table 1), the Hybrid GM-ROI model had the highest AUC and F1 scores. The second highest AUC score was observed for the ResNet3D (with 3D MRI Scans), which contrasts to its relatively poor performance with the accuracy metrics. Table 3 shows the comparison of unbalanced accuracies for different cutoff thresholds and confirms that the Hybrid GM-ROI model dominated the ResNet3D at all thresholds (see Table 3).

|  | I | T | All Patients | | Initial Severity: Severe or Moderate | |
| --- | --- | --- | --- | --- | --- | --- |
|  |  |  | AUC | F1 | AUC | F1 |
| Logistic | X | Y | 0.872 | 0.890 | 0.837 | 0.806 |
| Stitched MRI w/ ResNet-18 | Y | X | 0.873 +/- 0.002 | 0.868 +/- 0.002 | 0.819 +/- 0.003 | 0.752 +/- 0.006 |
| Stitched MRI w/ Lightweight CNN | Y | X | 0.861 +/- 0.001 | 0.872 +/- 0.002 | 0.807 +/- 0.002 | 0.750 +/- 0.005 |
| MRI Scans w/ ResNet3D | Y | X | 0.880 +/- 0.003 | 0.864 +/- 0.004 | 0.835 +/- 0.007 | 0.753 +/- 0.010 |
| Early Fusion Hybrid w/ ResNet-18 | Y | Y | 0.867 +/- 0.001 | 0.867 +/- 0.002 | 0.808 +/- 0.003 | 0.741 +/- 0.005 |
| DAFT | Y | Y | 0.879 +/- 0.003 | 0.861 +/- 0.004 | 0.820 +/- 0.008 | 0.741 +/- 0.013 |
| Hybrid GM-ROIs w/ ResNet-18 | Y | Y | **0.899 +/- 0.001** | **0.901 +/- 0.002** | **0.864 +/- 0.002** | **0.820 +/- 0.003** |
| Hybrid WM-ROI w/ ResNet-18 | Y | Y | 0.885 +/- 0.002 | 0.884 +/- 0.002 | 0.845 +/- 0.003 | 0.780 +/- 0.004 |
| Hybrid Stitched MRI w/ ResNet-18 | Y | Y | 0.887 +/- 0.002 | 0.879 +/- 0.003 | 0.841 +/- 0.003 | 0.768 +/- 0.005 |
| Hybrid Stitched Image w/ Lightweight CNN | Y | Y | 0.878 +/- 0.002 | 0.872 +/- 0.003 | 0.830 +/- 0.003 | 0.759 +/- 0.005 |
| GM-ROIs w/ ResNet-18 | Y | X | 0.876 +/- 0.001 | 0.877 +/- 0.002 | 0.846 +/- 0.002 | 0.777 +/- 0.006 |
| WM-ROI w/ ResNet-18 | Y | X | 0.868 +/- 0.002 | 0.864 +/- 0.002 | 0.795 +/- 0.003 | 0.745 +/- 0.005 |

Table 2. Area under the ROC curve and F1-scores for the test dataset. (I = uses image data, T = uses tabular data)

| Threshold | 0.1 | 0.2 | 0.3 | 0.4 | 0.5 | 0.6 | 0.7 | 0.8 | 0.9 |
|---|---|---|---|---|---|---|---|---|---|
| **Hybrid ROI** | 0.818 | 0.846 | 0.855 | 0.863 | 0.866 | 0.858 | 0.827 | 0.766 | 0.604 |
| **ResNet3D** | 0.808 | 0.825 | 0.828 | 0.827 | 0.818 | 0.799 | 0.749 | 0.658 | 0.497 |

Table 3. Comparison of unbalanced accuracy for different cutoff thresholds (confidence intervals are not shown for ease of reading).

Some additional analyses were carried out with the Hybrid ROI model, in order to understand its relatively strong performance. First, test runs were conducted to assess the contributions of its individual tabular features. New image datasets were created displaying the seven GM-ROIs plus either one or two of the tabular features. As shown in Table 4, 'initial severity' was found to have the largest impact, whilst 'left lesion size' had a negligible or negative impact. Hybrid RM-ROI images that only included 'initial severity' and 'recovery time' achieved highest accuracies in this paper; however, this may be the result of overfitting on the test dataset, as the choice of features was not selected using cross-validation. The apparent negative impact of 'left lesion size' when included with the other two features may be due to its signal already being present in the other features (ROIs, 'initial severity' and 'recovery time'), and that adding 'left lesion size' added noise that impaired classification. The effect of varying the depth of the ResNet architecture was also tested and it was found that increasing the depth slightly reduced the accuracies (see Table 5).

|  | All Patients | | Initial Severity: Severe or Moderate | |
|---|---|---|---|---|
|  | **Accuracy** | **Balanced accuracy** | **Accuracy** | **Balanced accuracy** |
| No features added | 0.832 +/- 0.003 | 0.811 +/- 0.004 | 0.764 +/- 0.008 | 0.763 +/- 0.008 |
| Initial severity | 0.851 +/- 0.002 | 0.841 +/- 0.003 | 0.799 +/- 0.006 | 0.799 +/- 0.006 |
| Left lesion size | 0.837 +/- 0.002 | 0.819 +/- 0.003 | 0.771 +/- 0.005 | 0.771 +/- 0.005 |
| Recovery time | 0.844 +/- 0.003 | 0.826 +/- 0.003 | 0.775 +/- 0.006 | 0.774 +/- 0.006 |
| Recovery time & Left lesion size | 0.845 +/- 0.001 | 0.829 +/- 0.002 | 0.785 +/- 0.003 | 0.785 +/- 0.003 |
| Initial severity & Left lesion size | 0.860 +/- 0.002 | 0.844 +/- 0.003 | 0.803 +/- 0.006 | 0.803 +/- 0.006 |
| Initial severity & Recovery time | **0.872 +/- 0.003** | **0.866 +/- 0.004** | **0.825 +/- 0.023** | **0.825 +/- 0.006** |
| All three features added | 0.866 +/- 0.002 | 0.855 +/- 0.002 | 0.822 +/- 0.004 | 0.822 +/- 0.004 |

Table 4. Accuracy results for modified versions of the Hybrid GM-ROI images. For example, 'Initial severity & Left lesion size' refers to experiments carried out with a dataset of images each displaying seven GM-ROIs plus the symbols representing the initial severity and left lesion size features, but without recovery time.

|  | ResNet-18 | ResNet34 | ResNet50 | ResNet101 |
|---|---|---|---|---|
| Accuracy | **0.866 +/- 0.002** | 0.855 +/- 0.002 | 0.859 +/- 0.002 | 0.857 +/- 0.002 |
| Balanced Accuracy | **0.855 +/- 0.002** | 0.842 +/- 0.003 | 0.847 +/- 0.003 | 0.842 +/- 0.005 |

Table 5. Accuracy results for ResNet models of different depths, trained on Hybrid GM-ROI dataset. The number at the end of ResNet is the number of layers in the network. These are the four smallest Pytorch ResNet models for which ImageNet weights are available.

5. *Discussion and Future Work*

This paper has shown that CNNs can provide predictions for aphasia recovery with a balanced accuracy of approximately 0.85. The best performance came from using 2D Hybrid ROI images that combined grey matter ROIs with three tabular features (initial severity of aphasia after stroke, left hemisphere lesion size and Recovery time). Of these three features, left hemisphere lesion size was least important when damage to key anatomical regions of interest was incorporated. It may seem surprising that the 3D CNNs were outperformed by some of the 2D models. A key issue is likely to be the number of patients in the project dataset. The number of voxels/features in an MRI scan massively exceeds the number of patient and this may lead to the curse of dimensionality. It could be that far larger patient numbers are needed to adequately populate the high dimensional feature space. Problems are further exasperated by the large number of trainable parameters in standard 3D CNNs. The lightweight 3D ResNet in Wolf et al.'s (2022) DAFT implementation might mitigate against the trainable parameters problem, but risks losing some of the predictive power of deeper ResNet models.

2D CNNs trained on the Stitched MRI dataset had similar accuracies to the 3D Resnet trained on the 3D MRI scans. The stitched MRI contain less information than the 3D MRI scans, as they only display 64 MRI slices that are downsized to 256 x 256 images. Yet this loss of information appears to be offset by having a smaller feature space and having less trainable parameters.

Table 1's accuracy results point to the Hybrid GM-ROIs and GM-ROIs datasets having greater prognostic information than either their respective Stitched MRI or WM-ROI datasets. As Figure 2 illustrates, this is the case even when the number of GM-ROIs being displayed is only four. However, it cannot be inferred that grey matter is etiologically most relevant to aphasia recovery. There are several reasons for this including (i) lesions lead to correlated damage between GM-ROIs and WM-ROIs. ii) the GM-ROIs will include some white matter voxels.

A key difficulty with the application of machine learning in neuroimaging (and more broadly) is the potential for over-fitting to creep in un-noticed (Hosseini et al., 2020). The difficulty is reflected in the bias-variance dilemma (Kohavi & Wolpert, 1996) (and its "twin": the trade-off between type-I and type-II errors (Lieberman & Cunningham, 2009)), i.e. changes that increase classification accuracy have the potential to hinder generalisation, or in other words, efforts to reduce under-fitting, can increase over-fitting. This is essentially because some of the improvement in classification accuracy is due to finding pattern in noise, rather than in signal. Additionally, this problem is especially serious when datasets are small, which in machine learning terms, ours is. The problem is that, with small data, the effective signal-to-noise ratio is also small. However, we believe that we have been diligent in protecting ourselves against gross overfitting. For example, use of a lock-box, which is only opened once (Hosseini et al.,2020), suggests that our reported accuracies reliably reflect the out-of-sample effectiveness of our learning algorithms, given the data available to us.

There is a subtle issue that if the (out-of-sample) accuracies of multiple learning algorithms are quantified on the *same* lock-box, the choice of the best amongst these will be inflated by this multiple testing. Ideally, one would like to have two lock-boxes, one to determine the best algorithm and a second to determine its true out-of-sample accuracy. However, if you are choosing between a relatively small number of algorithms (we have 12), using a single lock-box is not likely to be a large inflation of accuracy. All this said, replication in a new dataset, preferably by a new research group, is the ultimate test of generalization. We await this assessment.

There appears to be significant potential for increasing the predictive accuracy of CNN models for aphasia recovery. For example, the PLORAS dataset is planned to include an additional 2000 stroke survivors by 2028. Increasing the dataset size might reduce some of the problems with the curse of dimensionality and the large number of trainable parameters. The larger datasets may also improve the 2D CNNs' ability to learn complex patterns in the data, reflecting the heterogeneous nature of lesion patterns generating a particular deficit.

There is considerable scope for further developing the Hybrid image approach. For example, additional tabular features could be included such as age, sex at birth, handedness and the duration and intensity of treatments. Nonlinear transformations could also be applied to some of the tabular features, with the new values being represented by changes in the corresponding symbols' intensities or sizes. Symbolic data could also be added to the 3D MRI scans. Finally, Hybrid images could be created that combine GM-ROIs, WM-ROIs, and tabular data.

One reason we are able to obtain relatively high accuracies is due to the pre-training of the ResNet on the ImageNet dataset. For example, using pre-training weights improved the balanced accuracy of the Hybrid ROIs from 0.823 to 0.855. However, this pre-training is not focused on images relevant to the learning problem being considered, i.e. it is not on brain-scans. Consequently, if a very large dataset of T1-weighted MRI scans (hopefully, of 100s of thousands) can be identified then it may be possible to provide a pre-training that tunes the convolution kernels to features more appropriate for classification from the brain-scans available from stroke patients. There are a number of ways in which a teacher signal can be obtained for this pre-training (Doersch, Gupta & Efros, 2015). For example, for the CNN to be pre-trained could become the encoder in an autoencoder architecture, with the input scans, or parts of them, also serving as teacher pattern (Pathak et al., 2016). Additionally, if suitable cognitive measures are available with the pre-training dataset, then they could be used as the teacher signal. If available, training to classify language abilities should tune the CNN kernels very appropriately for classifying stroke recovery.

Finally, PLORAS is now collecting longitudinal data from 90 aphasic stroke survivors, including both MRI scans and extended tabular data. Changes in the voxel intensities and tabular features may well be prognostically valuable. These changes could be incorporated into Hybrid images.

6. *Limitations*

A key limitation of this work is that it has been restricted to research quality MRI scanners. If CNNs are to be clinically employed, then they will need to achieve high levels of accuracy using images from hospital scanners, including CT images.

A general limitation of using CNNs is that the complexity of their calculations is beyond human capacities to understand. Yet in a clinical setting it seems essential to be able to understand and explain why particular predictions are being made. This highlights the need for explainable AI methods. There are many explainable AI methods available. This paper has used a bespoke version of CLEAR Image, other methods include Grad-CAM and LIME. Unfortunately, these methods often differ in their putative explanations of a CNN's classification probabilities, highlighting different regions as being important (Fong et al., (2019), White et al. (2023)). A priority for explainable AI methods then, is to show that their explanations are faithful, i.e. they correctly mimic the input-output behaviour of the AI classifier that they are meant to be explaining. CLEAR Image does provide some fidelity statistics unlike Grad-CAM or LIME. However, further work is needed to assess the fidelities of explainable AI methods with MRI data.

There are two potential criticisms in the scope of the paper's experiments that were judged to be of low risk. The first is that there may be a different CNN architecture (e.g. EfficientNet) or a transformer that would have produced better results. However, we are unaware of any papers that indicate that a different architecture would be expected to generate substantially improved results compared to using ResNet models. It could also be argued that the project should have used data augmentation to increase the size of the training datasets. However, we are training CNNs to discriminate between small changes in lesion sizes and locations on spatially normalised images. The usual data augmentation transformations of rotation, blurring and translation would have distorted the subtle patterns that the CNN needed to learn (see Wang et al., (2023) for a similar argument when using CNNs to detect patterns in MRI scans of Alzheimer patients).

7. *Conclusions*

Predicting recovery from post-stroke aphasia could enable targeted therapy. This paper has provided a state-of-the-art assessment of the effectiveness of deep learning for predicting the class of a patient's spoken picture description score (i.e. aphasic/non-aphasic). We have shown that deep learning, multimodal data and feature selection using explainable AI can achieve high levels of predictive accuracy. However, if deep learning methods are to be clinically employed, then they will need to achieve high levels of accuracy using images from hospital scanners. There appears to be significant potential for achieving this, for example by increasing dataset size, developing the Hybrid image approach, better pre-training weights and using longitudinal data. Our findings may also be relevant to other neuroscience fields that wish to combine image data and tabular data. In cases where a dataset is small in machine learning terms, our novel approach of training a CNN on images that combine ROIs with symbolic representations of tabular data may be fruitful.

*Author Contribution*

Adam White -  Conceptualization, Methodology, Software, Formal Analysis, Writing – original draft.

Howard Bowman – Conceptualization, Methodology, Writing – review & editing, Supervision.

Margarita Saranti –– Conceptualization, Methodology, Writing – review & editing

Thomas M. H. Hope –– Conceptualization, Methodology, Writing – review & editing

Cathy J. Price – Conceptualization, Resources, Data Curation, Writing -review & editing, Supervision, Funding Acquisition.

Artur d'Avila Garcez – Methodology, Validation, Writing -review & editing

*Acknowledgments*

Data acquisition was funded by the Wellcome [203147/Z/16/Z;  205103/Z/16/Z; 224562/Z/21/Z], the Medical Research Council [MR/M023672/1] and the Stroke Association [TSA 2014/02]. PLORAS team members contributed to the acquisition and analysis of behavioural data. They include: Storm Anderson, Rachel Bruce, Megan Docksey, Kate Ledingham, Louise Lim, Sophie Roberts, and Hayley Woodgate. We are indebted to the patients and their carers for their generous assistance with our research. Margarita Saranti is supported by a Stroke Association Doctoral Fellowship (SA PGF 22\100013). For the purpose of Open Access, the author has applied a CC BY public copyright licence to any Author Accepted Manuscript version arising from this submission.

*Disclosure statement*

Authors declare they have no conflict of interest.

*Data/Code Availability*

The code for which will be made available on GitHub upon publication. The data used in this study are stored on the PLORAS database.